# An ADRC-Incorporated Stochastic Gradient Descent Algorithm for Latent Factor Analysis

Jinli Li, Ye Yuan, *Member, IEEE*

*Abstract*—High-dimensional and incomplete (HDI) matrix contains many complex interactions between numerous nodes. A stochastic gradient descent (SGD)-based latent factor analysis (LFA) model is remarkably effective in extracting valuable information from an HDI matrix. However, such a model commonly encounters the problem of slow convergence because a standard SGD algorithm only considers the current learning error to compute the stochastic gradient without considering the historical and future state of the learning error. To address this critical issue, this paper innovatively proposes an ADRC-incorporated SGD (ADS) algorithm by refining the instance learning error by considering the historical and future state by following the principle of an ADRC controller. With it, an ADS-based LFA model is further achieved for fast and accurate latent factor analysis on an HDI matrix. Empirical studies on two HDI datasets demonstrate that the proposed model outperforms the state-of-the-art LFA models in terms of computational efficiency and accuracy for predicting the missing data of an HDI matrix.

## I. Introduction

With the rapid development of big data technology, many industry applications are gradually turning to data science. Such applications usually contain numerous objects called nodes, i.e., users in social APPs [26], road sensors in intelligent transportation systems [3], and uses-items in recommender systems [7, 10]. Actually, it is impossible to acquire full interactions among them [19, 55, 57-60]. For example, in a recommender system, users only touch a tiny subset of items, but more interactive information is unknown [2, 11]. For describing this incomplete information, a high-dimensional and incomplete (HDI) matrix is used to represent the node's interactions. Note that the dimension of the HDI matrix represents the numerous nodes, and the matrix's incomplete entries stand for the interactive information. Consequently, the HDI matrix contains much fewer known entries (the known interactions) than the unknown entries (the unknown interactions) [13-15, 42].

Although most of the entries in HDI matrix are unknown, HDI matrix still contains plentiful valuable entries regarding various patterns [18, 19, 32-35]. Hence, it is important to research how to dig out needed information from known entries. In recent research, a latent factor analysis (LFA) model is a highly representative approach because of its high efficiency and expandability [1, 2, 4-6, 28-31, 38, 39]. An LFA model constructs two low-dimensional LF matrices and trains them based on the known entries, to make the desired LF matrices become a low-rank approximation to the target HDI matrix [8, 9, 41, 44-46].

Existing researches on LFA models [12-15] demonstrate that a standard stochastic gradient descent (SGD) algorithm has high scalability and efficiency in explaining an LFA model [7, 20, 48-54]. However, the SGD algorithm commonly encounters the situation of slow convergence because it acquires a latent factor depending on the stochastic gradient of current instance error only considering past update information [12]. In order to address this issue, it is more reasonable to improve the SGD's convergence rate further by considering all known historical information [56].

In the control fields, classic feedback controllers can reach the control target quickly by utilizing its all-historical information [36, 37, 40, 47]. Especially, one of the most successful controllers in feedback controller is a proportional-integral-derivative (PID) controller [27]. The PID controller utilizes historical information to compute the proportion, integral, and derivation terms, making the current updates becomes more comprehensive and reasonable. Therefore, study [16, 25] takes the learning error and training iteration of SGD-LFA model as the control error and time note in PID, and establishes the connection between machine learning and PID. For instance, Li et al. [16] utilize a nonlinear PID controller to reconstruct the learning errors by incorporating past instance errors, thereby improving a resultant model's convergence rate.

Such studies all utilize a standard PID controller to incorporate the past update information [43]. However, it has some drawbacks. For example, the operation of derivative and integral are not reasonable. Thus, Han et al. propose the Active disturbance rejection control (ADRC) to deal with this question [21]. It uses a more reasonable way to extract integral terms and derivative terms, and ensures the robustness of the model. Enlightened by this research, we imitate the process of accelerating the LFA model by PID, and propose an ADRC-incorporated SGD Latent Factor (ADSL) model by rebuilding the instance learning error by following the principle of an ADRC controller effectively.

Main contributions of this work include the following perspectives:

---

✧ J. L. Li is with the School of Computer Science and Technology, Chongqing University of Posts and Telecommunications, Chongqing 400065, China, and also with the Chongqing Key Laboratory of Big Data and Intelligent Computing, Chongqing Engineering Research Center of Big Data Application for Smart Cities, and Chongqing Institute of Green and Intelligent Technology, Chinese Academy of Sciences, Chongqing 400714, China (e-mail: appleli_li@163.com).
✧ Y. Yuan is with the College of Computer and Information Science, Southwest University, Chongqing 400715, China (e-mail: yuanyekl@gmail.com,).

a) **An ADS-based LFA model.** It incorporates an ADS algorithm to calculate a refined instant learning error, thereby improving the convergence rate of LFA on an HDI matrix.
b) **Experimental results on two HDI matrices.** It firstly proves that the proposed ADS-based LFA model significantly outperforms the state-of-the-art LFA models in terms of computational efficiency.

The rest part of this paper is organized as follows: Section II describes the preliminaries, Section III describes the FPS-based LFA model, Section IV describes the experiment and comparison of results, and finally, there are conclusions about this study in Section V.

## II. PRELIMINARIES

### A. Problem Formulation

As our fundamental data source, an HDI matrix is defined as [26-28]:

**Definition 1 (An HDI matrix):** Given two large node sets $M$ and $N$ denote two different sets of nodes respectively, each value $r_{m,n}$ in matrix $R \in R^{|M| \times |N|}$ denotes the node $m$'s preference on node $n$. Subset $\Lambda$ and $\Gamma$ denote the known and unknown entity sets of R. Note that if $|\Lambda| \ll |\Gamma|$, $R$ is an HDI matrix [15]

**Definition 2 (An LFA Model):** Given LF matrices $X \in R^{|M| \times f}$ and $Y \in R^{|N| \times f}$ denote the LF matrices of $M$ and $N$ respectively; $f$ denotes the dimension of LF matrices $X$ and $Y$, and $f \ll \min\{|M|, |N|\}$. Hence, an LFA model establishes a low-rank estimation $\hat{R}=XY^T$ defined on these two LF matrices and the known subset $\Lambda$ [16].

For acquiring the desired LF matrices, the loss function with Euclidean distance is formulated as follows [17]:

$$\varepsilon(X,Y) \triangleq \left( (R-\hat{R})^2 + \lambda \left( \|X\|_F^2 + \|Y\|_F^2 \right) \right)$$
$$= \sum_{r_{m,n} \in \Lambda} \left( (r_{m,n} - \hat{r}_{m,n})^2 + \lambda \|x_m\|_2^2 + \lambda \|y_n\|_2^2 \right); \quad (1)$$

where $\varepsilon$ denotes the learning objective, $r_{m,n}$ and $\hat{r}_{m,n}$ are elements in $R$ and $\hat{R}$, $x_m$ and $y_n$ are $m$-th and $n$-th row vectors of $X$ and $Y$, $\lambda$ denotes regularization coefficient [8, 19], $\|\cdot\|_F$ denotes Frobenius norm, and $\|\cdot\|_2$ computes the $L_2$ norm of a vector, respectively

### B. An SGD Algorithm for LFA

An SGD algorithm is commonly applied to optimize an LFA model [18]. The optimization of the objective function (1) with the SGD algorithm is shown as follows:

$$\arg\min_{X,Y} \varepsilon \overset{SGD}{\Rightarrow} \forall r_{m,n} \in \Lambda : \begin{cases} x_m \leftarrow x_m - \eta \dfrac{\partial \varepsilon_{m,n}}{\partial x_m}, \\ y_n \leftarrow y_n - \eta \dfrac{\partial \varepsilon_{m,n}}{\partial y_n}; \end{cases} \quad (2)$$

where $\eta$ denotes learning rate, $\partial \varepsilon_{m,n}/\partial x_m$ and $\partial \varepsilon_{m,n}/\partial y_n$ denote the gradients of $x_m$ and $y_n$, respectively. And the instant objective $\varepsilon_{m,n}$ is given as follows:

$$\varepsilon_{m,n} = \left( r_{m,n} - \langle x_m, y_n \rangle \right)^2 + \lambda \|x_m\|_2^2 + \lambda \|y_n\|_2^2. \quad (3)$$

Note that the instant learning error $e_{m,n} = r_{m,n} - \langle x_m, y_n \rangle$ corresponding to each instance $r_{m,n} \in \Lambda$. According to formulas (2) and (3), the SGD-based LFA model's learning is shown as follows:

$$\arg\min_{X,Y} \varepsilon \overset{SGD}{\Rightarrow} \forall r_{m,n} \in \Lambda : \begin{cases} x_m \leftarrow x_m + \eta \cdot (e_{m,n} \cdot y_n - \lambda \cdot x_m), \\ y_n \leftarrow y_n + \eta \cdot (e_{m,n} \cdot x_m - \lambda \cdot y_n). \end{cases} \quad (4)$$

### C. An ADRC Controller

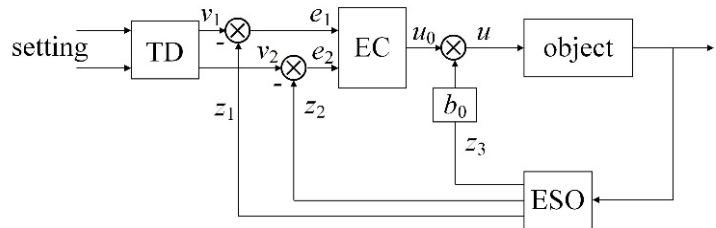

Fig. 1. Flowchart of ADRC.

The Active disturbance rejection control (ADRC), proposed by Han, has shown strong capabilities to handle external disturbances and internal uncertainties [21]. After more than ten years of development, it has matured in technology and expanded in the scope of application. An ADRC consists of three parts, i.e., tracking differentiator (TD), extended state observer (ESO), and error compensator (EC) [22]. The purpose of TD is to extract the input value and its differential value; the ESO estimates the real-time effect value of the internal and external disturbance of the system and gives compensation in the feedback, it eliminates the disturbance by compensation method, so it makes the ADRC can be anti-interference; the EC performs control and disturbance compensation based on the tracking value derived from TD and the error observed by ESO. The structure of an ADRC is shown in Fig. 1.

The process of ADRC is shown in Fig. 1. Mathematically, the ADRC is formulated as follows

$$\begin{aligned} e_1 &= v_1 - z_1, \\ e_2 &= v_2 - z_2, \\ u_0 &= b_1 \cdot e_1 + b_2 \cdot e_2, \\ u &= (u_0 - z_3)/b_0; \end{aligned} \quad (5)$$

where $v_1$ and $v_2$ denote the input value and its differential value derived from TD; $z_1$, $z_2$, and $z_3$ denote the output value, its differential value, and disturbance from ESO; $e_1$ and $e_2$ denote error and its differential; $b_1$ and $b_2$ denote the coefficients of $e_1$ and $e_2$; $b_0$ denotes the coefficient of compensation; $u_0$ denotes the error variation; $u$ denotes the error variation after compensation.

III. METHODS

*A. The ADRC-based Instant Learning Error Refinement*

Note that ADRC aims to exploit the feedback error to make the prediction value move towards the actual value [22]. Analogously, an SGD-based LFA model utilizes the SGD algorithm to reduce the learning error between actual and prediction values to achieve the goal of optimizing the objective function [16]. Therefore, it is reasonable to take the learning error as the feedback error and convert the SGD-based LFA model to an ADRC-based learning process. Moreover, the unique mechanism of ADRC makes it process several benefits, e.g. efficiency, flexibility, and robustness [23]. Hence, we plan to incorporate the ADRC into the SGD-based learning process to improve the LFA model's performance further.

Next, we will introduce the components of ADRC in turn to elaborate the process for refining the instant learning error.

**TD**: previous studies [16, 25] have successfully applied PID and non-linear PID to control the learning error of SGD-LFA and achieved promising results. However, the above models need to use the differential information of the instant learning error, the discrete PID controller just uses the difference between two adjacent errors to extract the differential of the instant error, so that the inaccurate derivative term doesn't play its due role in the improved models. Hence, ADRC provides a more precise way to acquire the differential value of instant learning error. In particular, ADRC applies TD and ESO to track the differential of true value and prediction value respectively, then, it uses the traced differential to do the difference to acquire the differential of the instant learning error.

Note that the learning rate $\eta$ overcomes the overshoot phenomenon of the LFA model to a great extent. Thus, our model does not need to acquire the true value and prediction value by TD and ESO. To train our ADSL, we compare time nodes in ADRC to iteration rounds in the SGD-LFA model, and the formula of discrete TD is shown as follows:

$$\begin{aligned} l &= -r \cdot \mathbf{sgn}\left(v_1(t-1) - \chi + \frac{v_2(t-1) \cdot |v_2(t-1)|}{2 \cdot r}\right), \\ v_1(t) &= v_1(t-1) + h \cdot v_2(t-1), \\ v_2(t) &= v_2(t-1) + h \cdot l; \end{aligned} \quad (6)$$

where $v_1(t)$ denotes the $t$-th tracking true value, $v_2(t)$ denotes the $t$-th tracking differential of true value, $\chi$ denotes the true value, $h$ denotes the integration step, and $r$ denotes the acceleration, respectively.

**ESO**: As the most important component in ADRC, ESO plays an essential role in the model's training. In detail, ESO observes the disturbance present in the model and cancels it out in advance, helping the model to obtain more accurate results. Meanwhile, ESO observes the differential of the predicted value and combines it with the differential of the true value tracked by TD, applying the error of the differential instead of the differential of the error to make the differential result more reasonable.

Note that our predicted values are passed directly back to the model, so only the differentials of the ESO observations are used. The formula of the discrete ESO is shown as follows:

$$\begin{aligned} e_1 &= z_1(t-1) - \hat{r}(t), \\ z_1(t) &= z_1(t-1) + h \cdot (z_2(t-1) - \beta_1 \cdot e_1), \\ z_2(t) &= z_2(t-1) + h \cdot (z_3(t-1) - \beta_2 \cdot e_1 + b \cdot u(t)), \\ z_3(t) &= z_3(t-1) - h \cdot \beta_3 \cdot e_1; \end{aligned} \quad (7)$$

where $z_1(t)$ denotes the $t$-th observed value of the predicted value, $z_2(t)$ denotes the $t$-th differential observation of the predicted value, $z_3(t)$ denotes the $t$-th observed value of disturbance, $\hat{r}(t)$ denotes the $t$-th predicted value, $b$ denotes the coefficient of compensation, $\beta_1$, $\beta_2$, and $\beta_3$ denote the parameters of the observer, $u(t)$ denotes the $t$-th control value.

**EC**: The purpose of EC is to calculate the refinement learning error based on the results of TD and ESO. The formula of the discrete EC is shown as follows:

$$\begin{aligned}
e_1(t) &= r - \hat{r}(t), \\
e_2(t) &= v_2(t) - z_2(t), \\
u_0(t) &= b_1 \cdot e_1(t) + b_2 \cdot e_2(t), \\
u(t) &= (u_0(t) - z_3(t))/b_0;
\end{aligned} \tag{8}$$

where $e_1(t)$ denotes the $t$-th instant learning error, $e_2(t)$ denotes the $t$-th differential of the instant learning error, $u_0(t)$ denotes the $t$-th refinement learning error before compensating the disturbance, $u(t)$ denotes the refinement learning error after compensating the disturbance, $b_0$, $b_1$, $b_2$ denote the gain parameters.

*B. The ADS Algorithm*

Considering the $t$-th iterations of an SGD-based LFA model, the formula (4) can be rewritten as follows:

$$\text{During the } t\text{-th training iteration: } \arg\min_{X,Y} \varepsilon \overset{SGD}{\Rightarrow}$$

$$\forall r_{m,n} \in \Lambda : \begin{cases} x_m \leftarrow x_m + \eta \cdot \left(e_{m,n}^{(t)} \cdot y_n - \lambda \cdot x_m\right), \\ y_n \leftarrow y_n + \eta \cdot \left(e_{m,n}^{(t)} \cdot x_m - \lambda \cdot y_n\right); \end{cases} \tag{9}$$

where $e_{m,n}^{(t)}$ denotes the instant learning error of instance $r_{m,n}$ at $t$-th training iteration. With the training rounds $t$ increase, the SGD algorithm repeats the formula (6) to optimize the LFA model according to the learning error until convergence.

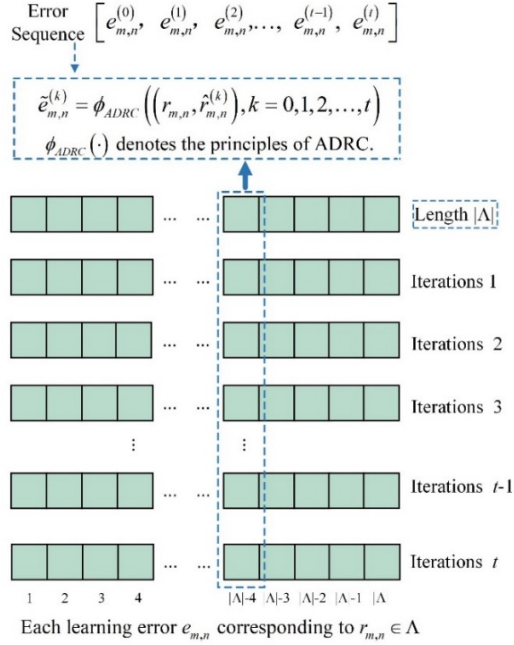

Fig. 2. Principle of learning error refinement.

From formula (9), we find that an SGD-based LFA model updates a single latent factor based on the instant learning error of each instance $r_{m,n} \in \Lambda$. Hence, we execute the ADRC-based instant learning error refinement for each instance $r_{m,n} \in \Lambda$ to obtain the corresponding refinement learning error.

That is, each instance is controlled by an independent ADRC controller. The flowchart of learning error refinement is shown in Fig. 2.

As shown in Fig. 2, we substitute the time point in ADRC with the training iterations in an SGD-based LFA model, and $|\Lambda|$ number of ADRC controllers are adopted to refine the instant learning error. Therefore, based on Section A, an error refinement principle for each instance's learning error is formulated as follows:

$$\begin{aligned}
l &= -r\,\mathbf{sgn}\!\left(v_{m,n}^{1(t-1)} - r_{m,n} + \frac{v_{m,n}^{2(t-1)} \cdot \left|v_{m,n}^{2(t-1)}\right|}{2 \cdot r}\right), \\
v_{m,n}^{1(t)} &= v_{m,n}^{1(t-1)} + h \cdot v_{m,n}^{2(t-1)}, \\
v_{m,n}^{2(t)} &= v_{m,n}^{2(t-1)} + h \cdot l;
\end{aligned} \tag{10}$$

where $v_{m,n}^{1(t)}$ and $v_{m,n}^{2(t)}$ denote the instance $r_{m,n}$'s tracking true value and its differential in $t$-th iteration, respectively.

$$e_{m,n}^{1(t)} = z_{m,n}^{1(t-1)} - \hat{r}_{m,n}^{(t)}, z_{m,n}^{1(t)} = z_{m,n}^{1(t-1)} + h \cdot \left( z_{m,n}^{2(t-1)} - \beta_1 \cdot e_{m,n}^{1(t)} \right),$$
$$z_{m,n}^{2(t)} = z_{m,n}^{2(t)} + h \cdot \left( z_{m,n}^{3(t-1)} - \beta_2 \cdot e_{m,n}^{1(t)} + b \cdot u_{m,n}^{(t)} \right), z_{m,n}^{3(t)} = z_{m,n}^{3(t-1)} - h \cdot \beta_3 \cdot e_{m,n}^{1(t)}; \quad (11)$$

where $z_{m,n}^{1(t)}$ and $z_{m,n}^{2(t)}$ denote the instance $r_{m,n}$'s observed predicted value and its differential in $t$-th iteration, $z_{m,n}^{3(t)}$ denote the $t$-th observed value of disturbance corresponding to instance $r_{m,n}$, respectively.

$$e_{m,n}^{(t)} = r_{m,n} - \hat{r}_{m,n}^{(t)}, e_{m,n}^{2(t)} = v_{m,n}^{2(t)} - z_{m,n}^{2(t)},$$
$$\tilde{e}_{m,n}^{0(t)} = b_1 \cdot e_{m,n}^{(t)} + b_2 \cdot e_{m,n}^{2(t)}, \tilde{e}_{m,n}^{(t)} = \left( u_{m,n}^{0(t)} - z_{m,n}^{3(t)} \right) / b_0; \quad (12)$$

Based on the above analysis, we obtain the ADRC-incorporated SGD-based (ADS) learning scheme as follows:

During the $t$-th training iteration: $\arg\min_{X,Y} \varepsilon \overset{ADSL}{\Rightarrow}$

$$\forall r_{m,n} \in \Lambda : \begin{cases} x_m \leftarrow x_m + \eta \cdot \left( \tilde{e}_{m,n}^{(t)} \cdot y_n - \lambda \cdot x_m \right), \\ y_n \leftarrow y_n + \eta \cdot \left( \tilde{e}_{m,n}^{(t)} \cdot x_m - \lambda \cdot y_n \right). \end{cases} \quad (13)$$

## IV. EXPERIMENTAL RESULTS AND ANALYSIS

### A. General Settings

**Evaluation Protocol.** In order to measure the accuracy of the ADS-based LFA model, this study chose the root mean squared error (RMSE) as the evaluation metric [16]. The lower RMSE value represents the higher prediction accuracy of the desired model. The formula of RMSE is shown as follows:

$$RMSE = \sqrt{\left( \sum_{r_{m,n} \in \Phi} \left( r_{m,n} - \hat{r}_{m,n} \right)^2 \right) / |\Phi|}$$

where $\hat{r}_{m,n}$ denotes the prediction value to $r_{m,n}$ generated on test data, $\Phi$ denotes the testing set, respectively. Meanwhile, we record the total training time of each model to evaluate the efficiency of the model. Note that all experiments are carried out on the same PC with a 3.2 GHz i5 CPU and 16 GB RAM.

**Experimental Datasets.** The details of two industrial HDI matrices adopted in our experiments are shown as follows:

TABLE I. Experimental dataset details.

| No. | Name | Row | Column | Known Entries | Density |
|---|---|---|---|---|---|
| D1 | ML10M | 71,567 | 10,681 | 10,000,54 | 0.31% |
| D3 | Douban | 129,490 | 58,541 | 16,830,839 | 0.22% |

In our experiment process, we use 70% of all known entries for each HDI data as the training set, 20% data rest as the testing set, and the remaining data as the validation set. Moreover, we use five-fold cross-validation in the above process. The conditions of training termination of each model are: a) the training iterations count exceeds 1000; or b) the RMSE difference between two consecutive iterations is smaller than $10^{-5}$.

### B. Comparison against State-of-the-art Models

We execute the experiment with several state-of-the-art LFA models. Table II gives the details of all the compared models. Table III gives the detailed performance of M1-4 on D1 and D2. Figs. 3 depicts the training curves of M1-4. From the above results, we acquire the following findings:

TABLE II. Details of compared models.

| Model | Description |
|---|---|
| M1 | The proposed ADS-based LFA model of this study. |
| M2 | A standard PID-incorporated SGD-based LFA model [25]. |
| M3 | An LFA model with a PID-based stochastic optimizer [24]. |
| M4 | A standard SGD-based LFA model [12]. |

TABLE III. Lowest RMSE and their corresponding total time cost (Secs).

| Case | | M1 | M2 | M3 | M4 |
|---|---|---|---|---|---|
| D1 | RMSE: | **0.7927** | 0.7929 | 0.7948 | 0.7939 |
| | Time: | **39.6(s)** | 71.4(s) | 79.7(s) | 134.4(s) |
| D2 | RMSE: | **0.7251** | 0.7252 | 0.7290 | 0.7257 |
| | Time: | **90.0(s)** | 150.1(s) | 167.5(s) | 336.8(s) |

a) **The proposed ADSL model acquires superior computational efficiency gain over its peers.** For example, as shown in Table III, on D1, M1 takes 39.6 seconds to converge in RMSE, which is about 44% (i.e., ($Cost_{high}$-$Cost_{low}$)/ $Cost_{high}$) lower than M2's 71.4 seconds, 50% lower than M3's 79.7 seconds, 70% lower than M4's 134.4 seconds. The same outcomes are also encountered in D2, as recorded in Table III.

b) **The ADSL model also acquires better prediction accuracy for missing data.** More specifically, as depicted in Table III, on D1, M1 achieves the lowest RMSE at 0.7927, which is about 0.02% lower than 0.7929 by M2, 0.26% lower than 0.7948 by M3, 0.15% lower than 0.7939 by M4. In the other testing cases, similar results are encountered as shown in Table III.

## V. Conclusions

This paper proposes an ADS-based LFA model for fast and accurate latent factor analysis on an HDI matrix. It incorporates the ADRC controller into an LFA algorithm for acquiring better convergence performance. In addition, extensive experimental results demonstrate that ADS-based LFA is superior both in computational efficiency and prediction accuracy for missing data of an HDI matrix.